\documentclass[runningheads]{llncs}

\usepackage[mobile]{eccv}

\usepackage{eccvabbrv}

\usepackage{graphicx}
\usepackage{booktabs}
\usepackage{multirow}
\usepackage{epigraph}
\usepackage{pifont}

\usepackage[accsupp]{axessibility}  % Improves PDF readability for those with disabilities.

\usepackage{hyperref}

\usepackage{orcidlink}

\begin{document}

\title{Vamos: Versatile Action Models for Video Understanding} 

\titlerunning{Vamos}


\author{
Shijie Wang$^{1}$
\qquad
Qi Zhao$^{1}$
\qquad
Minh Quan Do$^{1}$
\qquad
Nakul Agarwal$^{2}$\\
Kwonjoon Lee$^{2}$
\qquad
Chen Sun$^{1}$
}

\authorrunning{S.~Wang et al.}

 \institute{$^{1}$Brown University \quad $^{2}$Honda Research Institute USA}
 
\maketitle
\begin{abstract}

What makes good representations for video understanding, such as anticipating future activities, or answering video-conditioned questions? While earlier approaches focus on end-to-end learning directly from video pixels, we propose to revisit text-based representations, such as general-purpose video captions, which are interpretable and can be directly consumed by large language models (LLMs). Intuitively, different video understanding tasks may require representations that are complementary and at different granularity. To this end, we propose versatile action models (Vamos), a learning framework powered by a large language model as the ``reasoner'', and can flexibly leverage visual embedding and free-form text descriptions as its input. To interpret the important text evidence for question answering, we generalize the concept bottleneck model to work with tokens and nonlinear models, which uses hard attention to select a small subset of tokens from the free-form text as inputs to the LLM reasoner. We evaluate Vamos on five complementary benchmarks, Ego4D, NeXT-QA, IntentQA, Spacewalk-18, and EgoSchema, on its capability to model temporal dynamics, encode visual history, and perform reasoning. Surprisingly, we observe that text-based representations consistently achieve competitive performance on all benchmarks, and that visual embeddings provide marginal or no performance improvement, demonstrating the effectiveness of text-based video representation in the LLM era. We also  demonstrate that our token bottleneck model is able to select relevant evidence from free-form text, support test-time intervention, and achieves nearly 5 times inference speedup while keeping a competitive question answering performance. Code and models are publicly released at \href{https://brown-palm.github.io/Vamos/}{https://brown-palm.github.io/Vamos/}.

\end{abstract}    
\section{Introduction}

Building a generative model for everyday human activities has long been a dream for researchers working on video understanding. Central to this problem are capturing the interactions between humans and the environment~\cite{wang2016actions,damen2018scaling}, modeling the temporal dynamics of activities~\cite{vondrick2016generating,epstein2021learning}, and encoding the hierarchical structures among atomic actions~\cite{sadanand2012action,gu2018ava}, activities~\cite{caba2015activitynet,carreira2017quo}, and events~\cite{ke2007event,jiang2013high}. Once constructed, the generative model of actions
can be applied to a wide range of tasks, including activity and event recognition~\cite{pastra2012minimalist}, future behavior prediction~\cite{liu2020forecasting}, goal and intent inference~\cite{pei2011parsing}, and temporal reasoning~\cite{xiao2021next}.

Despite its desirable properties, generative modeling of actions from video observations remains challenging, hindered by two open research questions: First, what makes good video representations? Earlier attempts often relied on manually defining the actions and the objects being interacted with~\cite{gu2018ava,aein2019library,sadanand2012action}. They require task-specific prior knowledge, and cannot generalize to the ``open vocabulary'' scenarios in the wild. Alternative approaches aim to model the temporal dynamics of human pose~\cite{minderer2019unsupervised,li2021ai,kalakonda2022action} or latent representations encoded by deep neural networks~\cite{vondrick2016generating,jayaraman2018time}, which are either too fine-grained, or not directly intepretable. Second, what makes a good model of human actions? While earlier approaches attempted to apply rule-based generative action grammars~\cite{ivanov2000recognition,hongeng2004video,nevatia2004ontology,pastra2012minimalist}, they may not be able to capture the diverse, even peculiar ways of how events would unfold over time. More recent approaches adopt a data-driven framework and directly learn autoregressive models~\cite{generative_action} on visual tokens~\cite{sun2019videobert,yan2021videogpt}, where the visual domain is often specialized (\eg cooking, or robotics).

\begin{figure}[t]
    \centering
    \includegraphics[width=\textwidth]{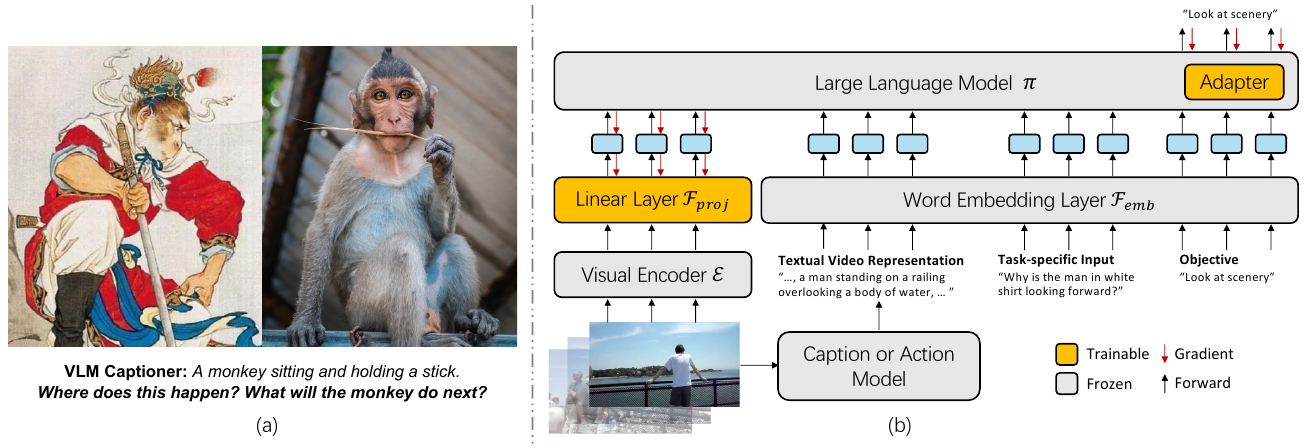}
    \captionsetup{width=0.9\textwidth} 
    \caption{(a) Visual observations with vastly diverse appearances may be described with the same captions. Our work explores the use of general-purpose text descriptions of video data for action anticipation and question answering. We propose Vamos, a versatile reasoning framework that allows us to study the impact of latent visual representations and free-form text descriptions for downstream applications. 
    (b) Overview of the Vamos framework. It flexibly unifies distributed visual features and text-based representations such as video captions, and can be applied to diverse video understanding tasks.}
    \label{fig:model}
\end{figure}

We aim to address both challenges by exploring an unconventional idea: Can \textbf{task-agnostic} natural language descriptions, such as those generated by off-the-shelf image caption models~\cite{li2023blip,liu2023visual} on sampled video frames, serve as useful video representations for action modeling from videos? And if so, can we then leverage a pre-trained large language model (LLM)~\cite{touvron2023llama} as the generative model of actions, represented as free-form text?  If both answers are yes, visual reasoning can then enjoy recent advances in LLM research, where they have been shown to be capable of learning context-free grammar~\cite{allen2023physics} with long-range dependencies, predicting time-series~\cite{gruver2023large,mirchandani2023large}, and performing reasoning~\cite{boix2023can,dasgupta2022language,wei2022chain,ding2021attention}, all of which are indispensable for action modeling. Since we assume the text descriptions are general purpose, they can be extracted once and reused for different downstream tasks, similar to pre-computed visual embeddings~\cite{radford2021learning}.

To rigorously validate this idea, we propose versatile action models (\textbf{Vamos}), a framework that flexibly unifies three representations, namely distributed visual embeddings and free-form text descriptions,
and can be  applied to various applications by leveraging 
large language models, such as Llama-2~\cite{touvron2023llama}. The visual embeddings are linearly projected into the same language space following standard practice~\cite{li2023blip,moon2023anymal}. 
As illustrated in Figure~\ref{fig:model}(a), the same caption can sometimes be used to describe visually diverse inputs, it is thus essential for Vamos to be able to leverage one or multiple representations simultaneously, to understand the impact of individual representation type.
Vamos can directly leverage an LLM's next token prediction capability for action anticipation~\cite{grauman2022ego4d}. We also ask Vamos to perform video question answering~\cite{xiao2021next}, by appending the question to the video representation as inputs to the LLM reasoner.

One inherent benefit of text-based video representation is its interpretability. Inspired by interpretable object classifiers such as concept bottleneck models (CBM)~\cite{koh2020concept}, we aim to understand which words serve as important evidence for question answering. However, CBM requires a pre-defined list of discrete visual concepts, and requires a linear classifier to achieve interpretability. We generalize this framework and learns hard attention to select a small subset of text inputs to feed to the LLM reasoner, where the text inputs are tokenized text as opposed to pre-defined concepts. We call our generalized formulation token bottleneck models (TBM). TBM naturally supports the incorporation of multimodal information, and allows users to perform causal intervention.

We perform extensive evaluations on five benchmarks, including the Ego4D dataset~\cite{grauman2022ego4d} for long-term action anticipation, NeXT-QA~\cite{xiao2021next} and  IntentQA~\cite{li2023intentqa} for video question answering, Spacewalk-18~\cite{krishnan2023spacewalk} for long-form procedural video understanding, and EgoSchema~\cite{mangalam2023egoschema} for zero-shot long-form video question answering. We observe that for the direct application of Vamos in the action anticipation task, the text-based representation outperforms its counterpart based on visual embeddings. We further observe that free-form video descriptors serve as an effective long-video representation that generalizes well in zero-shot setting, outperforming the strongest video-language model~\cite{wang2022internvideo} by 66\%. We then confirm that our observations are general, that text-based representation consistently provides competitive performance across all tasks, and that adding visual embeddings surprisingly results in marginal or no performance gains. Finally, we demonstrate that the token bottleneck model is able to select semantically relevant evidence for question answering, and achieves 5x speedup at inference time while maintaining the question answering accuracy.

\iffalse
To summarize, we make the following contributions:

\begin{enumerate}
\item We propose to revisit video representation and action modeling in the LLM era, and to explore the effectiveness of free-form text descriptions.
\item We propose Vamos, a versatile framework that allows us to incorporate and compare the effectiveness of different video representations.
\item We demonstrate the effectiveness of free-form text as a representation that is performant, interpretable, and can be intervened. Vamos achieves state-of-the-art performance on Ego4D LTA, NeXT-QA, IntentQA, and EgoSchema.
\end{enumerate}
\fi
\section{Related Work}

\noindent \textbf{Vision-Language Foundation Models.} Models such as CLIP~\cite{radford2021learning} and ALIGN~\cite{jia2021scaling} bridge the vision and language modalities by learning a text encoder and an image encoder jointly with a contrastive loss on image and caption pairs.
Another line of vision-language models~\cite{li2019visualbert,chen2020uniter,li2020oscar,singh2022flava} combines the masked language modeling objective with image-text contrastive learning, and focus on downstream tasks such as visual question answering, visual commonsense reasoning, and text-guided object detection. For videos, VIOLET~\cite{fu2021violet} trains an end-to-end transformer for video and language modalities, by representing videos as visual tokens and performing joint masked token modeling. To perform visual-language joint training, speech transcripts are often used as the language modality for videos~\cite{sun2019videobert,fu2021violet,zellers2021merlot,zellers2022merlot}.
The objectives can be combined~\cite{yu2022coca} and the encoders for different modalities can be shared~\cite{wang2022allinone}. Compared to existing VLMs, Vamos imposes an ``information bottleneck'' when text-based representation is used: It converts visual inputs into discrete action labels and free-form text descriptions.

\noindent\textbf{Visually-augmented LLMs.} Apart from joint visual-language pre-training, existing large language models (LLMs) can also be augmented to incorporate visual inputs. For example, VisualGPT~\cite{chen2022visualgpt} and Flamingo~\cite{alayrac2022flamingo} directly fuse visual information into the layers of a language model decoder using a cross-attention mechanism instead of using images as additional prefixes to the language model. Other approaches, such as instructional tuning~\cite{liu2023visual}, prompting large language models for knowledge retrieval~\cite{shao2023prompting}, or linearly projecting the visual embeddings into the input space of LLMs~\cite{moon2023anymal, lin2023vila}, have also been explored. Vamos largely follows this approach to incorporate visual embeddings, with the goal to understand if and how they are complementary to text-based video representations.

Additionally, tool-using large language models have been recently proposed to invoke and incorporate the use of task-specific modules~\cite{schick2023toolformer,hu2023avis}, where visual perceptions consist a substantial subset of the tools. Notably, VisualProgram~\cite{gupta2023visual} and ViperGPT~\cite{suris2023vipergpt} propose to apply LLMs to generate symbolic programs based on pre-selected computer vision modules for visual question answering. VidIL~\cite{wang2022language} leverages expert knowledge to design object and action concepts for few-shot captioning and video question answering. 
Closest to our work is Socratic Models~\cite{zeng2022socratic}, where the authors propose to use natural language as the common interface to connect foundation models with different input and output modalities. Finally, several concurrent works~\cite{zhang2023simple, li2023mvbench, min2024morevqa} share similar motivations and methods to Vamos in utilizing text-based representations and modules for video understanding, without the incorporation of visual embeddings.

\section{Method}
We now describe how the text-based representation is constructed and incorporated into versatile action models.

\subsection{Text-based Video Representation}

A video often contains complex and dynamic information including context and interactions. While prior works~\cite{koh2020concept,yuksekgonul2022post,wei2023diffusion} have demonstrated the effectiveness of condensing images into text-based representations such as visual concepts, it remains unclear if videos can also be condensed into text-based representations. To answer this research question, we consider text descriptions that are task-agnostic, and can potentially be applied in diverse video understanding tasks. 

Concretely, we rely on general-purpose captioning models to generate free-form text descriptions to characterize objects, scenes, and actions, which succinctly summarize the essential elements depicted in the video.
We employ off-the-shelf image captioning models such as BLIP-2~\cite{li2023blip2} that generate image-level captions from the sampled video frames. These captions are subsequently concatenated to form a comprehensive video-level caption.

For certain tasks, prior knowledge might be helpful to guide the model learning. For example, when the goal is to model the long-term temporal dynamics of verbs and nouns for the long-term action anticipation task, it would be beneficial to trim the inputs to only contain discrete action labels. 
In practice, this can be achieved through the application of action recognition models such as Transformer encoders that operate in the pre-defined action space. 

\subsection{Versatile Action Models}
Large language models (LLMs) have demonstrated strong capability for temporal reasoning~\cite{zhao2023antgpt} and even some potential for causal reasoning~\cite{kiciman2023causal}, both of which are crucial for video understanding. We introduce Vamos, a simple yet effective framework to utilize LLMs to unify video dynamic modeling tasks, including comprehending historical content (video question answering, VQA) and future prediction (long-term action anticipation, LTA). As shown in Figure~\ref{fig:model} (b), given a video $V$ and a pretrained LLM $\pi$, the input sequence $\mathbf{x}_t=[\mathbf{x}_\text{tvr}, \mathbf{x}_\text{task}]$ consists of the textual video representations $\mathbf{x}_\text{tvr}$ of $V$ and other task specific language inputs $\mathbf{x}_\text{task}$ (e.g., instructions, questions, targets). The frozen word embedding layer $\mathcal{F}_\text{emb}$ first generate the corresponding text tokens $\mathbf{z}_t = \mathcal{F}_\text{emb}(\mathbf{x}_t) \in \mathbb{R}^{L_t\times D}$, where $L_t$ is the sequence length of $\mathbf{x}_t$, $D$ is the feature dimension.

Vamos incorporates the \textit{residual} information not entirely captured by $\mathbf{x}_\text{tvr}$ via representations encoded directly from the visual modality, such as CLIP visual embedding. We adopt a learnable linear projection layer $\mathcal{F}_\text{proj}$ to align visual features with the language space. Specifically, the frozen vision backbone $\mathcal{E}$ takes in $N_v$ frames $[v_1,...v_{N_v}]$ sampled from $V$ to generate the visual features. These visual features are then fed into the projection layer $\mathcal{F}_\text{proj}$ to produce visual tokens $\mathbf{z}_{v} = \mathcal{F}_\text{proj}(\mathcal{E}(v_1,...v_{N_v}))\in \mathbb{R}^{N_v\times D}$. To combine information from the visual and textual representations, we adopt the early fusion strategy and concatenate $\mathbf{z}_v$ and $\mathbf{z}_t$ as the inputs to the LLM $\pi$. When labeled training data is available for task-specific fine-tuning, we update the weights of the LLM $\pi$ either with LoRA~\cite{hu2021lora} or LLaMA-Adapter~\cite{zhang2023llama}.

Vamos can accommodate diverse video understanding tasks by formulating each task as sequence completion given an appropriate task description $\mathbf{x}_\text{task}$, the LLM $\pi$ can then be optimized with the standard language modeling objectives. Specifically, for the VQA task, $\mathbf{x}_\text{task}$ is composed of instructions, questions, and answers, with the answer being the training objective. During inference, the answer that maximizes sequence modeling likelihoods is selected for multiple-choice QA, or directly generated for open-ended QA. For the LTA task, $\mathbf{x}_\text{task}$ is composed of instructions and future actions, where the training objective is the future action sequence. During inference, the fine-tuned LLM is tasked to generate sequences of future actions based on the history actions.

\noindent\textbf{Discussion:} By design, Vamos naturally incorporates temporal information as the captions are timestamped. Since the text-based representation is general-purpose, the framework is efficient since once extracted, the text-based representation can be reused for different questions, just like the CLIP embeddings commonly used by VLMs. Vamos is a two-stage framework with decoupled ``perception'' and ``reasoning'' modules, in addition to the benefits on interpretability (of the intermediate representation) and efficiency (reuse of the intermediate features). Another conceptual benefit is its generalizability: While the visual distributions of different datasets may differ, the reasoning module may be shared. We show this is indeed the case for EgoSchema~\cite{mangalam2023egoschema}, when the end-to-end vision-language models performance significantly worse than Vamos.

\subsection{Token Bottleneck Models}

\begin{figure}[t]
    \centering
    \includegraphics[width=\textwidth]{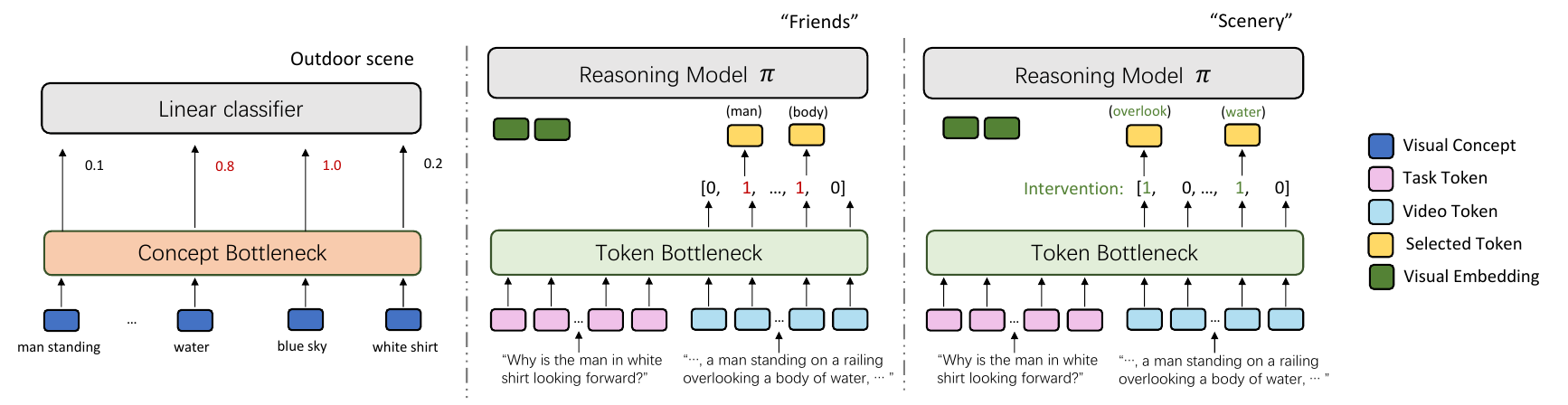}
    \captionsetup{width=0.9\textwidth} 
    \caption{An illustration of the token bottleneck model (TBM). We are inspired by the concept bottleneck models (CBM)~\cite{koh2020concept}, which achieve interpretable object classification by inspecting the weights of the learned linear classifier (left). Unlike CBM, Vamos does not require pre-defining a list of concepts. It directly works with tokenized text inputs. To provide input tokens to the reasoning model (an LLM), we leverage hard attention to generate binary rather than continuous weights (middle). The token bottleneck can be interpreted directly. It can also be intervened with human inputs (right), or augmented with residual visual information.}
    \label{fig:token_bottle}

\end{figure}

We aim to understand how the text-based representations are utilized by the LLM reasoners. As an LLM operates like a black box, we aim to enhance the interpretability of the overall framework by first understanding how it selects evidence to solve the downstream tasks. As illustrated in Figure~\ref{fig:token_bottle}, we are inspired by the success of interpretable object classifiers, such as the concept bottleneck model (CBM)~\cite{koh2020concept}. CBM relies on pre-defining a list of concepts and building (often supervised) concept detectors for each of them. As our model's inputs are free-form text, we propose to directly work with word tokens as opposed to pre-defined concepts. In addition, CBM relies on linear classifiers to achieve model interpretability. Each weight in a learned classifier indicates the importance of the corresponding concept for making the prediction. We hypothesize that linear classifiers are not sufficiently expressive when solving tasks that require a stronger ``reasoner''. To strike a balance between model interpretability and expressiveness, we generalize the CBM framework to learn binary attention on the input tokens as opposed to continuous weights used by the CBM linear classifier. As illustrated in Figure~\ref{fig:token_bottle}, The binary weights indicate which tokens are to be selected and fed to the more expressive LLM for solving the target tasks. We name this generalized framework as token bottleneck models (TBMs).

To implement TBM, we design a lightweight token selector as an add-on module for Vamos (Figure~\ref{fig:token_bottle} middle). It takes the tokenized embeddings for the text-based video representations $\mathbf{z}_\text{tvr}$ and the task-specific tokens $\mathbf{z}_\text{task}$ as its inputs. It learns to pick a single token among the candidate tokens $\mathbf{z}_\text{tvr}$ by optimizing the objective for a given downstream task. To select a sequence of tokens, we assume the important information is even distributed across the input sequence, and uniformly divide the input sequence into $k$ segments $\{\mathbf{z}_\text{tvr}^{(1)}$,..., $\mathbf{z}_\text{tvr}^{(k)}\}$, each of which contains $n$ tokens. Each segment $\mathbf{z}_\text{tvr}^{(i)} = \{z_1^{(i)}, ..., z_n^{(i)}\}$ is fed into the token selector, from which one token $z^{(i)}$ is selected for the task $\mathbf{z}_\text{task}$.

Within the token selector, $\{z_1^{(i)}, ..., z_n^{(i)}\}$ are first projected to a lower dimension, and then provided as inputs to a shallow transformer encoder to obtain encodings $\{s_1^{(i)}, ..., s_n^{(i)}\}$. A linear layer then takes these encodings and generates the logits $\mathbf{g}^{(i)} \in \mathbb{R}^n$ for final selection. During training, we apply Gumbel-Softmax~\cite{jang2016categorical} on the logits $\mathbf{g}^{(i)}$ to pick the token $z^{(i)}$ for each segment $\mathbf{z}_\text{tvr}^{(i)}$ while ensuring the module is differentiable. In this way, $k$ tokens are selected as the condensed representation of the original tokenized input sequence $\mathbf{z}_\text{tvr}$.

The token selector in TBM allows us to inspect the important evidence selected for the downstream tasks, and to intervene the wrongly recognized or selected tokens with the correct ones with human in the loop (Figure~\ref{fig:token_bottle} right).
Practically, the token selector also can also speed up the inference time due to its own light-weight implementation, and that only a much smaller subset of the tokens (e.g. 6\%) are processed by the computationally heavy LLM. 
\section{Experiments}

\noindent
In this section, we conduct experiments on two tasks and four datasets with both quantitative and qualitative analysis.
\subsection{Task and Datasets}

\noindent \textbf{Long-term action anticipation.} The LTA task asks a model to predict a sequence of actions in form of a verb-noun pairs in a long future window based on video observations of the past actions. In LTA, a long video $V$ is first split into a number of annotated video segments. Given video observation before segment $i$, our task is to predict the future actions in sequences of verb-noun pairs of the next $Z$ segments allowing $K$ candidate sequences. The correctness of the predicted sequence is measured with edit distance. We evaluation on:

\textit{Ego4D}~\cite{grauman2022ego4d} is comprised of 3,670 hours of egocentric videos in hundreds of scenarios of daily life activity. The Ego4D LTA v2 benchmark we focus on includes a total duration of around 243 hours of videos annotated into 3472 clips with 117 verbs and 521 nouns. We follow the official dataset splits and adopt the official parameters of the evaluation metric, with $Z=20$ and $K=5$.

\noindent \textbf{Video question answering.} Given a set of videos $V$, and a corresponding set of language-based questions $Q_v$ and their candidate answers $A_q$. The goal of video question answering (VQA) task is to predict the correct answer $A$ for each video-question pair. The performance is measured by accuracy. For VQA, we evaluate on three datasets:

\textit{EgoSchema}~\cite{mangalam2023egoschema} is annotated on Ego4D videos for long-form video QA. Each video is around 3 minutes and the \textit{temporal certificate} for humans to solve each task is around 100 seconds. It has 5,031 videos and each video is annotated with a multiple-choice question. All examples are for zero-shot evaluation.

\textit{Spacewalk-18}~\cite{krishnan2023spacewalk} is a long-form procedural video understanding benchmark collected on 18 spacewalk videos. The total duration is over 96 hours. We evaluate on the step recognition task which has a temporal certificate of 140 seconds. We follow the zero-shot setup with 1-minute context window, and report step recognition accuracy on the test set.

\textit{NeXT-QA}~\cite{xiao2021next} is a popular multiple choice video question answering benchmark that tests video understanding in terms of describing and reasoning the temporal actions. It contains 5,440 video clips and 47,692 questions, grouped into causal (48\%), temporal (29\%) and descriptive (23\%). 

\textit{IntentQA}~\cite{li2023intentqa} is a multiple choice VQA dataset built on top of NeXT-QA but focuses on intent reasoning. The authors select the videos related to causal and temporal questions from NeXT-QA and constructed their own questions and answers to focus on testifying models's performance on reasoning questions.

\subsection{Implementation}
\noindent \textbf{Generating Action Labels.} To generate action labels for videos, we use a recognition model pretrained on Ego4D LTA. It is a 3-block 6-head transformer-encoder that takes 4 CLIP features and outputs two logits for verb and noun respectively. It predicts actions in the action space pre-defined by Ego4D LTA. For each video, our action recognition model samples 4 frames for each 8s segment splits uniformly and output a verb and noun pair for the segment.

\noindent \textbf{Generating Video Captions.} We generate zero-shot, task-agnostic video captions using BLIP-2 for Ego4D LTA, IntentQA, and EgoSchema. We use LLaVA-1.5 to generate captions for NeXT-QA and Spacewalk-18 due to its better performance (see Table~\ref{tab:ablation}). 
For Ego4D LTA, we sample the center frame for each video segment to generate its caption. The captions for the 8 observed segments are then concatenated as the video representations. For VQA benchmarks, we first uniformly sample a fixed number of frames for each video, and then generate and concatenate the frame-level captions. We sample 6, 6, 12, and 12 frames for NeXT-QA, IntentQA, Spacewalk-18, and EgoSchema, respectively. For Spacewalk-18, we additional use its provided speech narrations.

\noindent \textbf{Vamos for Temporal Modeling.}
For full-shot VQA and LTA, We use LLaMA-7B and LLaMA2-7B~\cite{touvron2023llama} respectively as the temporal model for video understanding. During training, we use LLaMA Adapter~\cite{zhang2023llama} or low-rank adaption~\cite{hu2021lora} (LoRA) to perform parameter-efficient fine-tuning on the training set. For vision input, we use the frozen CLIP~\cite{radford2021learning} ViT-L/14 to extract image features. For the zero-shot long-form VQA on EgoSchema, we use several popular LLMs including OpenAI GPT-3.5-turbo, GPT-4, GPT-4o and LLaMA2-chat-13B.

\begin{table}[h]
\centering
\captionof{table}{\textbf{Vamos with various video representations.} Results are reported on Ego4D LTA test set, the metric is edit distance. Results are reported on NeXT-QA validation set, and IntentQA test set, the evaluation metric is accuracy. On all datasets, text-based representations achieve competitive performance.}
\scalebox{1}{
\begin{tabular}{c|ccc|cccc|cccc}
\toprule
 \multirow{2}{*}{\textbf{Input}} & \multicolumn{3}{c|}{\textbf{Ego4D-LTA $\downarrow$}} & \multicolumn{4}{c|}{\textbf{NeXT-QA $\uparrow$}}  &  \multicolumn{4}{c}{\textbf{IntentQA  $\uparrow$}} \\
                       & Verb &Noun &Action & Cau.  & Tem.  & Des.  & All & CW  & CH  & TP\&TN  & All \\ 
                       \midrule
 vision                &0.653 &0.673 &0.884 &69.6 &67.2 &74.7 &69.6 &68.9 &71.6 &58.0 &66.7 \\
 text               & 0.661 &0.651 &0.878 &\textbf{75.5} &\textbf{71.3} &81.1 &\textbf{75.0} &\textbf{74.0} &\textbf{78.6} &\textbf{67.5} &\textbf{73.2} \\
 vis+text          & \textbf{0.643} &\textbf{0.650} &\textbf{0.868}  &74.5 &71.0 &\bf81.7 &74.5 &73.5 &76.6 &64.3 &71.7\\
\bottomrule
\end{tabular}
}
\label{tab:main_vqa}
\end{table}

\subsection{LLM as Long-term Video Temporal Reasoner}

We first apply Vamos on the long-term action anticipation task, which requires direct modeling of video temporal dynamics by predicting future actions tokens based on video observation. We fine-tune Vamos on Ego4D LTA dataset with continuous visual embeddings or text-based representation. We observe that action-based representation slightly outperforms the free-form captions due to the nature of the task. As shown in the first three columns of Table~\ref{tab:main_vqa}, we observe that the text-based representation outperforms the vision-based input, and combining the two further improves the performance. 

\noindent
\begin{minipage}{0.45\textwidth}
\centering
\captionof{table}{\textbf{zero-shot} VQA on Egoschema. *: 500 question subset.}
\scalebox{0.9}{
\begin{tabular}{lcr}
\toprule
Model & Input Type  & \qquad Acc.  \\ 
\midrule
InternVideo~\cite{wang2022internvideo} & frame & 32.1\%\\
GPT-4  & text & \textbf{48.26}\% \\
GPT-4* & gt-narration & 81.80\% \\
\bottomrule
\end{tabular}
}
\label{tab:egoschema}
\end{minipage}
\hfill
\begin{minipage}{.5\textwidth}
\centering
\captionof{table}{Ablation on the number of frames on EgoSchema with GPT-3.5 turbo.}
\scalebox{0.9}{
\begin{tabular}{cc}
\toprule
 \# Frames  &\qquad Full Set Acc.  \\ 
\midrule
1 & 37.83\%\\
4 & 38.36\%\\
12 & 41.24\% \\
\bottomrule
\end{tabular}
}
\label{tab:egoschema-frame}
\end{minipage}

We now consider the more challenging task, long-form video question answering, by conducting zero-shot experiment on the recently collected EgoSchema benchmark. We employ OpenAI GPT-4 as the video ``reasoner''. We observe that free-form captions extracted by BLIP-2 consistently outperform the action-based representation, presumably because richer information is retained in the captions. We report the free-form caption-based performance from now on, unless otherwise mentioned. From Table~\ref{tab:egoschema}, we observe that Vamos with text-based video representation largely outperforms the state-of-the-art InternVideo model~\cite{wang2022internvideo}, which is jointly trained on vision and language inputs. We attribute the performance gain due to the decoupling of perception and reasoning, the latter of which we hypothesize is easier to generalize even at zero-shot, thanks to the LLM pre-training.
To better understand the performance upper-bound of the text-based representation, we use 500 ground-truth video narrations provided by the authors of EgoSchema to perform evaluation on this subset. Remarkably, the LLM achieves an impressive accuracy of 81.8\% with the oracle captions. Although not directly comparable with the full-set performance, the result confirms the potential of the Vamos framework for reasoning over broad time spans in the long-form video question answering task, and that better empirical performance may be achieved by improving the captioning models.

\subsection{What Makes Good Video Representation?}
In addition to text features, Vamos can also integrate vision-language features. We then investigate whether different modalities encode complementary information on Ego4D, IntentQA, and NeXT-QA. We perform parameter efficient fine-tuning to update the weights of the LLM, whose gradients are used to jointly train the linear projection layer to incorporate the visual embeddings. We observe that naively fine-tuning with text and visual inputs lead to model overfitting, and hence perform modality dropout (i.e. randomly discard the entire sequence of visual embeddings) when fine-tuning the vis+text models.

While visual and text inputs are complementary in the Ego4D LTA task, we observe in Table~\ref{tab:main_vqa} that the caption-based representation significantly outperform the vision-only baseline in the NeXT-QA and IntentQA benchmarks for video question answering, and that adding visual features only marginally affects the performance. This suggests that pre-trained visual embeddings such as CLIP may not encode residual information useful for the video QA task (e.g. fine-grained visual and motion features). Extracting visual representations that would complement the information encoded by the task-agnostic captions is thus an important future work.

\begin{figure}
    \centering
\includegraphics[width=\textwidth]{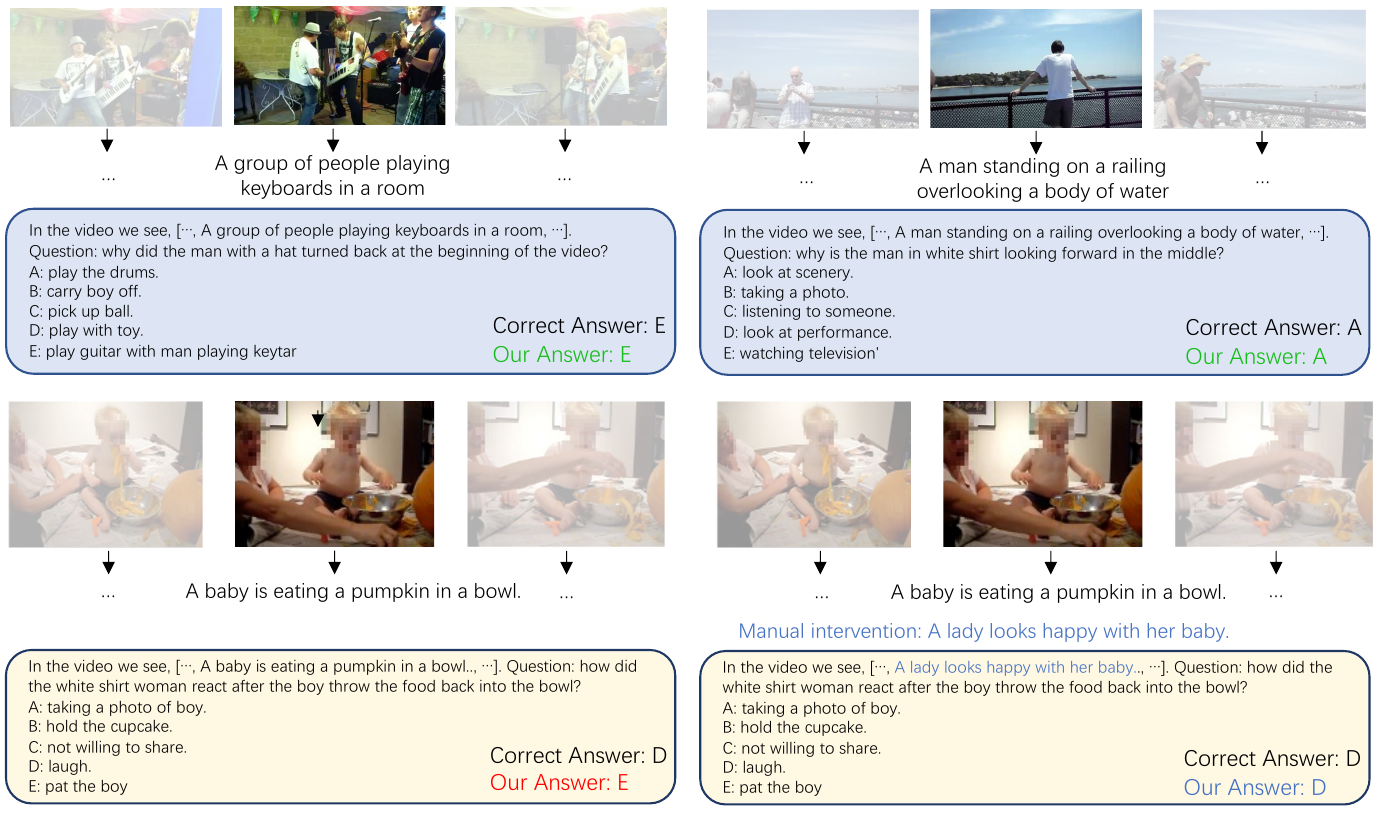}
 \captionsetup{width=\textwidth}
    \caption{Visualizations of example Vamos predictions and manual intervention.}
    \label{fig:iqavis}
\end{figure}

\subsection{Token Bottleneck Models}
Although the text-based video representation is directly interpretable, it remains unclear which tokens are selected as evidence for making predictions. The token bottleneck models (TBMs) enable us to reveal the selected evidence and also improve the model's inference speed.

\noindent \textbf{VQA Performance.} We
first tokenize and embed the input sequence $\mathbf{x}_t$ to obtain $\mathbf{z}_t$, which is uniformly partitioned into $k=$ 20 or 40 segments. TBM is then applied on each segment to select one token. In this way, we condense a long sequence (644 tokens on average for NeXT-QA) into $k$ tokens. TBM is jointly optimized with Vamos with LLaMA-3 during training.

For NeXT-QA dataset, we follow previous setting to use captions from 6 frames generated by LLaVA-1.5. For comparison, we also show the vision-only performance taking in 12 frames and the performance with unselected caption-based input. 
Table~\ref{tab:atp} shows that the TBM leads to an expected performance drop due to discarding over 90\% of the input tokens, and increasing $k$ improves the TBM performance. When $k=40$, Vamos achieves a competitive 69.6\% accuracy, while only adding 0.7M parameters to the Vamos framework and achieving \textbf{5x inference speedup} from 1.41s to 0.29s per sample on a single A6000 GPU.

\begin{figure}[h]
    \centering
    \includegraphics[width=.9\textwidth]
    {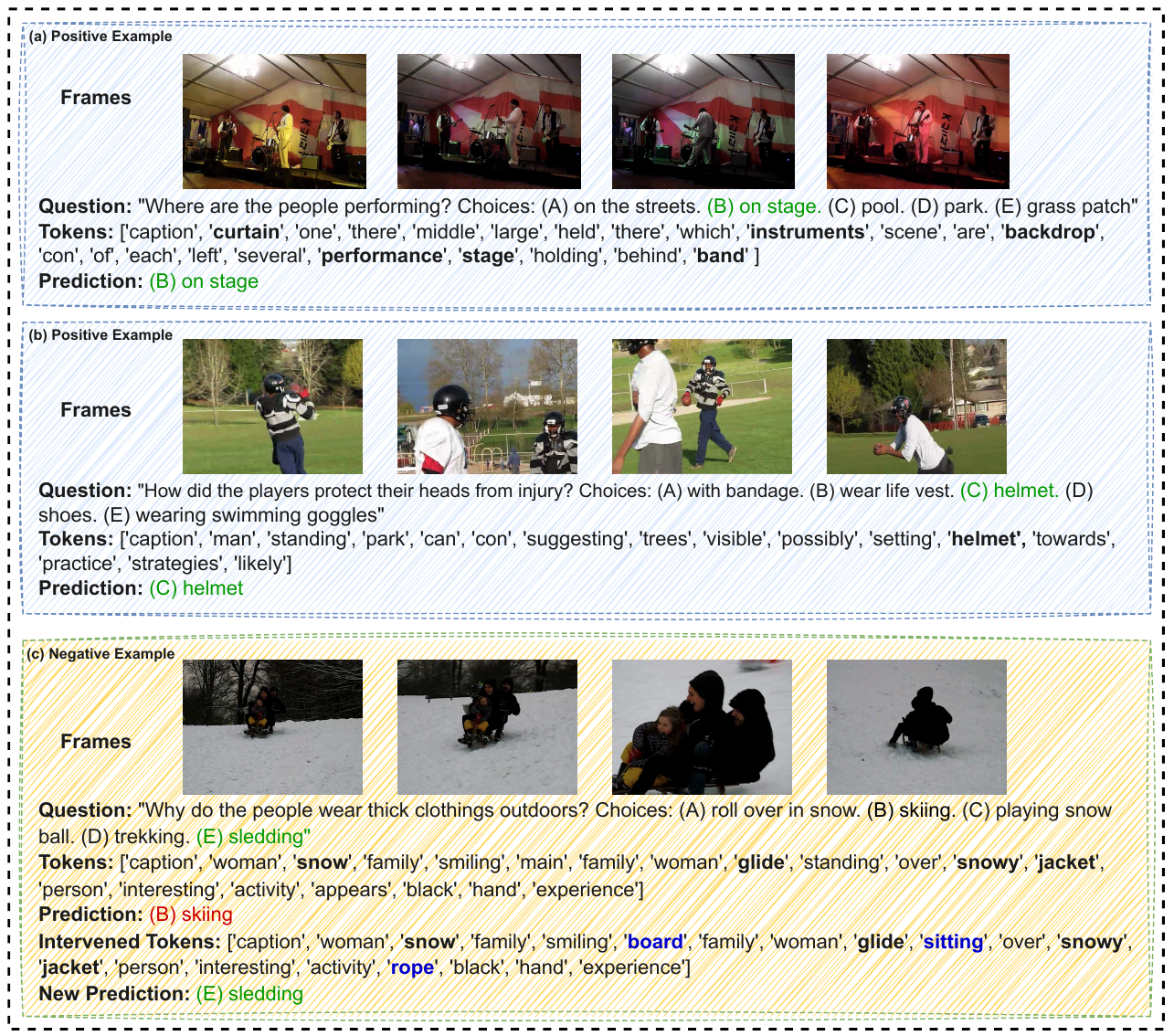}
    \captionsetup{width=\textwidth}
    \caption{Illustration of the token bottleneck model and the impact of intervention on model's predictions. The selected tokens are highly related to the task (examples $a$ and $b$) and can be manually intervened to correct the model's prediction (example $c$).}
    \label{fig:apt_vis}
\end{figure}

\begin{table}[h]
\centering
\captionof{table}{Condensing captions with token bottleneck models on NeXT-QA.}
\scalebox{1}{
\begin{tabular}{p{4em}p{10em}ccccc}
\toprule
Input & Selected tokens ($k$) & Cau. & Tem. & Des.  & All  \\ 
\midrule
vision  & all &71.9 &67.4 &75.6 &71.0\\
text  & all &77.2 &75.3 &81.7 &77.3\\
\midrule
text  & 20 / 644 & 68.1 & 64.6 & 70.8 & 67.4 \\
text  & 40 / 644 & 70.1 & 67.4 & 72.2 & 69.6 \\

\bottomrule
\end{tabular}
}
\label{tab:atp}
\end{table}

\noindent \textbf{Visualization of Vamos predictions.} Figure~\ref{fig:iqavis} provides two positive and one negative examples selected from IntentQA. The predictions are made by Vamos without TBM.
From the two positive examples we can see that the generated captions manage to describe the scene and activities happening in the video (``overlooking water'' and ``playing keyboards''), thus providing strong clues for LLMs to answer the question about description (``look at scenery'') and reasoning (``play guitar with man playing keytar''). However, a sub-optimal caption will also cause the Vamos's failure of reasoning. In the third example, we observe that the captions successfully describe the baby's action of eating but fail to describe the presence and potential actions of a woman in the corner of the scene. This omission leads to an incorrect prediction regarding the woman's reaction.

\noindent \textbf{Visualization of the Selected Tokens.} TBM not only helps accelerate Vamos' inference speed, but also allows us to interpret which pieces of evidence are used by the model to make predictions.
In Figure~\ref{fig:apt_vis}, we show two positive examples and one negative example of Vamos with TBM from NeXT-QA. The two positive examples demonstrate Vamos' ability to select tokens that serve as evidence for question answering. The negative example is challenging: even highly related tokens such as ``snow'' and ``glide'' are selected by Vamos, it still fails to reason the correct activities in the video.

\noindent \textbf{Test Time Intervention.} Text-based video representation is not only directly interpretable as used by Vamos and TBM, it also allows the users to conduct post-hoc, test-time intervention, which is pivotal for diagnosing and fixing failed predictions without retraining. In Figure~\ref{fig:iqavis} and Figure~\ref{fig:apt_vis}, we show examples of test-time intervention performed on the negative examples for Vamos and Vamos with TBM, respectively. By providing more accurate and related captions or tokens, Vamos is able to correct its wrong predictions.

\noindent
\begin{minipage}{.48\textwidth}
\captionof{table}{Ablation on captioning model and frame numbers on NeXT-QA.}
\scalebox{.8}{
\begin{tabular}{cccccc}
\toprule
\ Caption & \# Frame & Cau. & Tem. & Des.  & All  \\ 
\midrule
LLaVA-13B &1 &69.9 &67.2 &74.8 &69.8 \\
LLaVA-13B &3 & 73.0 & 70.4 & 79.4 & 73.1\\
LLaVA-13B &6 &\textbf{75.5} &71.3 &81.1 &75.0 \\
LLaVA-7B &6 &75.2 &\textbf{71.9} &\textbf{81.6} &\textbf{75.1} \\
BLIP-2 &6 &72.7 &68.9 &78.8 &72.4 \\
\bottomrule
\end{tabular}
}
\label{tab:ablation}
\end{minipage}
\hfill
\begin{minipage}{0.5\textwidth}
\centering
\captionof{table}{Comparison of different LLaMA models on NeXT-QA.}
\scalebox{1}{
\begin{tabular}{cccccc}
\toprule
Model  & Cau. & Tem. & Des.  & All  \\ 
\midrule
LLaMA1-7B & 75.5 &71.3 &81.1 &75.0  \\
LLaMA2-7B & 74.8 &72.3 &81.6 &75.0\\
LLaMA3-8B & \bf77.2 &\bf75.3 &\bf81.7 &\bf77.3 \\
\bottomrule
\end{tabular}
}
\label{tab:llama_ablation}
\end{minipage}

\subsection{Design Choices and Ablation Study}
\noindent \textbf{Caption Models.}
We study the impact of caption models on the video QA performance on NeXT-QA. We compare two captioning models: BLIP-2 and LLaVA-1.5. We observe that captions generated by BLIP-2 are generally more concise (less than 20 tokens), while captions generated by LLaVA-1.5 are more detailed (around 100 tokens on average). Results shown in Table~\ref{tab:ablation} shows that captions from LLaVA-1.5 achieve better performance. We also investigate the influence of caption model size by comparing LLaVA-1.5 7B and 13B versions. Interestingly, scaling LLaVA from 7B to 13B does not lead to improvement.

\noindent \textbf{Number of Frames.}
We study the impact of sampled frame numbers for captioning on NeXT-QA. As shown in Table~\ref{tab:ablation}, we found that using more captioned frames leads to better performance. A similar trend can be observed on EgoSchema in Table~\ref{tab:egoschema-frame}. However, we observe diminishing return when 12 frames are used, and hence not worth the speed and accuracy trade-off.

\noindent \textbf{Impact of LLMs on VideoQA.} 
We study the impact of different LLaMA versions on NeXT-QA. As shown in Table~\ref{tab:llama_ablation}, Vamos directly benefits from a more advanced LLM, which is a desirable property for practitioners.

\noindent \textbf{Impact of LLMs on Long-form VideoQA.}
We study the impact of LLMs on the EgoSchema benchmark. As shown in Table~\ref{tab:egoschema-sota}, GPT-4o achieves significant improvements comparing with GPT-4 and LLaMA2-Chat-7B baselines, which again demonstrates that Vamos can directly benefit from advances in LLMs.

\subsection{Comparison with State-of-the-art}
We compare our proposed Vamos with other state-of-the-art models in Tables~\ref{tab:sota_nextqa},~\ref{tab:sota_intent},~\ref{tab:sota_lta}, and~\ref{tab:egoschema-sota}. We train Vamos with LLaMA2-7B on Ego4D LTA and LLaMA3-8B on NeXT-QA and IntentQA. On EgoSchema zero-shot VQA, our approach based on GPT-4o outperforms the best vision-language model by 66.8\% on the full set.
On NeXT-QA, Vamos achieves the best performance and significantly outperforms LLaMA-VQA~\cite{ko2023large} and SeViLA\cite{yu2023self}, even though the former uses a LLM with 33B parameters and the latter is trained on additional dataset with temporal localization supervision. On IntentQA, Vamos also outperforms all baselines, with a 28.7\% accuracy improvement compared to the best performing prior method. On Ego4D LTA, Vamos outperforms previous works, without relying on domain-specific video encoders~\cite{lin2022egocentric}. Finally, on Spacewalk-18, we use GPT-4o as the LLM and achieves 18.6\% accuracy, which significantly outperforms the prior best zero-shot performance of 13.6\%.

\noindent
\begin{minipage}{.48\textwidth}
\centering
\captionof{table}{Comparison on NeXT-QA benchmark. * with additional supervision.}
\scalebox{0.7}{
\begin{tabular}{c cccc}
\toprule
\ Model & Cau. & Tem. & Des.  & All  \\ 
\midrule
ATP~\cite{buch2022revisiting} & 53.1\% & 50.2\% & 66.8\% & 54.3\% \\
HiTeA~\cite{ye2023hitea} & 62.4\% & 58.3\% & 75.6\% & 63.1\% \\
Intern Video~\cite{wang2022internvideo} & 62.5\% & 58.5\% & 75.8\% & 63.2\% \\
BLIP-2~\cite{li2023blip2} & 70.1\% & 65.2\% & 80.1\% & 70.1\% \\
SeViLA*\cite{yu2023self} & 74.2\% & 69.4\% & 81.3\% & 73.8\% \\
LLaMA-VQA-7B~\cite{ko2023large} & 72.7\% & 69.2\% & 75.8\% & 72.0\% \\
LLaMA-VQA-33B~\cite{ko2023large} & 76.2\% &72.6\% &78.8\% &75.5\% \\
\textbf{Vamos (ours)} & \textbf{77.2}\% &\textbf{75.3}\% &\textbf{81.7}\% &\textbf{77.3}\% \\
\bottomrule
\end{tabular}
}
\label{tab:sota_nextqa}
\end{minipage}
\hfill
\begin{minipage}{.48\textwidth}
\centering
\captionof{table}{Comparison with SOTA on IntentQA.}
\scalebox{0.75}{
\begin{tabular}{c cccc}
\toprule
\ Model & CW & CH & TP\&TN  & ALL  \\ 
\midrule
HGA~\cite{jiang2020reasoning} & 44.88\% & 50.97\% & 39.62\% & 44.61\% \\
HQGA~\cite{hqga} & 48.24\% & 54.32\% & 41.71\% & 47.66\% \\
VGT~\cite{xiao2022video} & 51.44\% & 55.99\% & 47.62\% & 51.27\% \\
BlindGPT~\cite{ouyang2022training} & 52.16\% & 61.28\% & 43.43\% & 51.55\% \\
CaVIR~\cite{li2023intentqa} & 58.4\% & 65.46\% & 50.48\%  & 57.64\% \\
\textbf{Vamos (ours)} & \textbf{75.14}\% & \textbf{77.44}\% & \textbf{69.58}\% & \textbf{74.16}\%\\
\bottomrule
\end{tabular}
}
\label{tab:sota_intent}
\end{minipage}

\noindent
\begin{minipage}{.4\textwidth}
\centering
\captionof{table}{Comparison with SOTA on Ego4D LTA v2 test set.}
\scalebox{0.9}{
\begin{tabular}{c cccc}
\toprule
\ Model & verb & noun & action \\ 
\midrule
Slowfast~\cite{feichtenhofer2019slowfast} & 0.717 & 0.736 & 0.925 \\
VideoLLM~\cite{chen2023videollm} & 0.721 & 0.725 & 0.921 \\
PaMsEgoAI~\cite{ishibashi2023technical} & 0.684 & 0.679 & 0.893 \\
Palm~\cite{huang2023palm} & 0.696 & 0.651 & 0.886 \\
AntGPT~\cite{zhao2023antgpt} &0.650  &\textbf{0.650}  &0.877 \\
\textbf{Vamos (ours)} &\textbf{0.643} &\textbf{0.650} &\textbf{0.868} \\
\bottomrule
\end{tabular}
}
\label{tab:sota_lta}
\end{minipage}
\hfill
\begin{minipage}{.55\textwidth}
\centering
\captionof{table}{Egoschema VQA zero-shot performance on full set and subset.}
\scalebox{0.8}{
\begin{tabular}{cccc}
\toprule
Model & Input Type  &  Full   &  Subset\\ 
\midrule
FrozenBiLM~\cite{yang2022zero} & frame & 26.9\% &-\\
mPLUG-Owl~\cite{ye2023mplug}   & frame & 31.1\% &-\\
InternVideo~\cite{wang2022internvideo} & frame & 32.1\% &-\\
\textbf{Vamos} (LLaMA2-13B) & caption & 36.73\% &38.20\%\\
\textbf{Vamos} (GPT-3.5) &caption & 41.24\% &47.60\%\\
\textbf{Vamos} (GPT-4)  & caption & 48.26\% & 51.20\%\\
\textbf{Vamos} (GPT-4o)  & caption & \textbf{53.55}\% &\textbf{57.20}\%\\

\bottomrule
\end{tabular}
}
\label{tab:egoschema-sota}
\end{minipage}

\iffalse
\begin{minipage}{.5\textwidth}
\centering
\captionof{table}{Comparison with SOTA on STAR}
\scalebox{0.9}{
\begin{tabular}{c ccccc}
\toprule
\ Model & Int. & Seq. & Pre. & Fea. & ToT. \\ 
\midrule
AIO~\cite{wang2023all} & 47.5 & 50.8 & 47.8 & 44.1 & 47.5 \\
ATP~\cite{buch2022revisiting} & 50.6 & 52.9 & 49.4 & 40.6 & 48.4 \\
MIST~\cite{gao2023mist} & 55.6 & 54.2 & 54.2 & 44.5 & 53.9 \\
InternVideo~\cite{wang2022internvideo} & 62.7 & 65.6 & 54.9 & 51.9 & 58.7 \\
\textbf{Vamos (ours)} &\textbf{} &\textbf{} &\textbf{} \\
\bottomrule
\end{tabular}
}
\label{tab:sota_lta}
\end{minipage}
\fi
\section{Conclusion}

We study different forms of video representations and propose versatile action models (Vamos) as a unified framework to utilize visual- and text-based representations for video understanding. We conduct extensive experiments on long-term action anticipation and video question answer benchmarks. Surprisingly, we observe that direct applications of free-form, general-purpose text-based video representations, such as captions, serve as strong video representation for all benchmarks we consider. Vamos utilizes large language models to perform zero-shot reasoning, and incorporate human feedback via test-time intervention. We further propose the token bottleneck models, which allow the users to interpret the evidence selected by Vamos, and speed up inference by nearly 5x. Vamos achieves state-of-the-art results on Ego4D LTA, IntentQA, NeXT-QA, Spacewalk-18, and outperforms the best vision-language model by over 66\% on EgoSchema. 

\noindent\textbf{Limitations:} Although our results show the promise of free-form text-based representations, we believe visual information is still essential for complex video understanding and reasoning. We expect future work to investigate alternative visual encoders~\cite{tong2024eyes} to extract fine-grained visual information beyond what has been captured by the captions, and to propose pre-training paradigms that better align visual inputs with the input space of LLMs. We also believe better benchmarks that require fine-grained, structured visual understanding are needed to rigorously evaluate the impact of representations for video understanding.

\section*{Acknowledgements}
This work is supported by Honda Research Institute USA and Samsung Advanced Institute of Technology. We would like to thank Karttikeya Mangalam, Raiymbek Akshulakov, and Shyamal Buch for their kind help with EgoSchema and ATP; Apoorv Khandelwal, Calvin Luo, David Isele, Songpo Li, and Tian Yun for their useful feedback and discussions. Our research was conducted using computational resources at the Center for Computation and Visualization at Brown University.

\bibliographystyle{splncs04}
\bibliography{main}

\newpage
\appendix
\setcounter{table}{0}
\renewcommand{\thetable}{A\arabic{table}}
\setcounter{figure}{0}
\renewcommand{\thefigure}{A\arabic{figure}}

\section{Additional Experiments and Results}

\noindent \textbf{Textual representation format.} On the Ego4D LTA task,
we observe that explicitly regularizing the text-based inputs to contain only verbs and nouns lead to slightly improved performance (0.878 versus 0.890 action edit distance). We hypothesize that this is due to the nature of the task, which focuses solely on predicting future verbs and nouns. In Table~\ref{tab:egoschema_type}, we again compare the action-based and caption-based text representations, but on the EgoSchema zero-shot VQA task. 
In contrast to the LTA performance, the caption-based representation performs much better, intuitively because solving the video question answering would require more details about the video, which can be provided by the general-purpose captions.

\noindent \textbf{Frame number on EgoSchema.} In Table 3 we compare different frame numbers and the zero-shot performance on EgoSchema. We uniformly sample 1, 4, and 12 frames from the videos and concatenate the captions of each frame to form the text video representations. We use GPT-3.5 as the reasoner. The results show that using more frames leads to better performance as more information and temporal evidence are provided.

\noindent \textbf{Temporal ordering information.} We investigate the influence of temporal information for long-form video understanding by shuffling the frame order when concatenating captions. In Table~\ref{tab:egoschema-shuffle}, we compare the ordered and shuffled 12-frame captions. Surprisingly, the performance drop (2\%) by shuffling the video captions is not as significant as we expect. We suspect that LLMs may have strong capability of ``auto-correcting'' the order of the input sentences, even if they are shuffled.

\begin{minipage}{0.45\textwidth}
\centering
\captionof{table}{Comparison of action and caption for \textbf{zero-shot} VQA on Egoschema.}
\scalebox{0.9}{
\begin{tabular}{c cc}
\toprule
Model & Input Type  & Full Set Acc.  \\ 
\midrule
GPT-4  & action & 38.12\% \\
GPT-4  & caption & \textbf{48.26}\% \\
\bottomrule
\end{tabular}
}
\label{tab:egoschema_type}
\end{minipage}
\hfill
\begin{minipage}{.45\textwidth}
\centering
\captionof{table}{Ablation on the influence of temporal information. 12 frames are sampled.}
\scalebox{0.9}{
\begin{tabular}{cc}
\toprule
 Shuffle  &\qquad Full Set Acc.  \\ 
\midrule
\ding{52} & 39.22\%\\
 & \bf{41.24}\% \\
\bottomrule
\end{tabular}
}
\label{tab:egoschema-shuffle}
\end{minipage}

\section{Addition Implementation Details}
\noindent \textbf{Vamos.}
We train Vamos on 4 A6000 GPUs for 2, 5, 10 epochs on Ego4D, IntentQA, and NeXT-QA respectively. For NeXT-QA, we set the maximum sequence length of pure vision input as 128, and 1200 for captions and captions + vision inputs, for IntentQA, the sequence lengths are set as 128 and 512 respectively.

\noindent \textbf{Token bottleneck model.}
For the token bottleneck model (TBM), we use a 2-layer transformer encoder with 2 attention heads, and a hidden size of 128. No additional positional encoding is added by the token bottleneck model. In order to condense the input sequence for interpretability and select the most relevant tokens, the TBM is \textit{task dependent}, namely taking the questions and the candidate answers (when available) as inputs along with the video captions. The selected tokens, as opposed to their corresponding encoded embeddings by the token bottleneck, are fed into the LLM, as illustrated in Figure 2 (middle) in the main paper.

\section{Additional Visualization}
Figure~\ref{fig:taskvisshort} illustrates the input format for long-term action anticipation and video question answering tasks. In the LTA task, the task-specific inputs are the discrete action labels and the target output is a future action sequence. For supervised VQA, the task-specific input is composed of instructions, video representations (vision and text), question and choices. The target output is the chosen answer to the question. For zero-shot VQA on EgoSchema, the video representation only consists of the text descriptions. Figure~\ref{fig:taskvisshort} shows the prompt designs for all tasks.

Figure~\ref{fig:cap_vis} shows four example video captions generated by LLaVA and BLIP-2 on the NeXT-QA dataset, respectively. We observe that in general the LLaVA-generated captions are longer and more detailed. Recall that Table 5 shows that the more detailed LLaVA captions also lead to higher VQA performance.

\begin{figure*}[h]
    \centering
    \includegraphics[width=\textwidth]{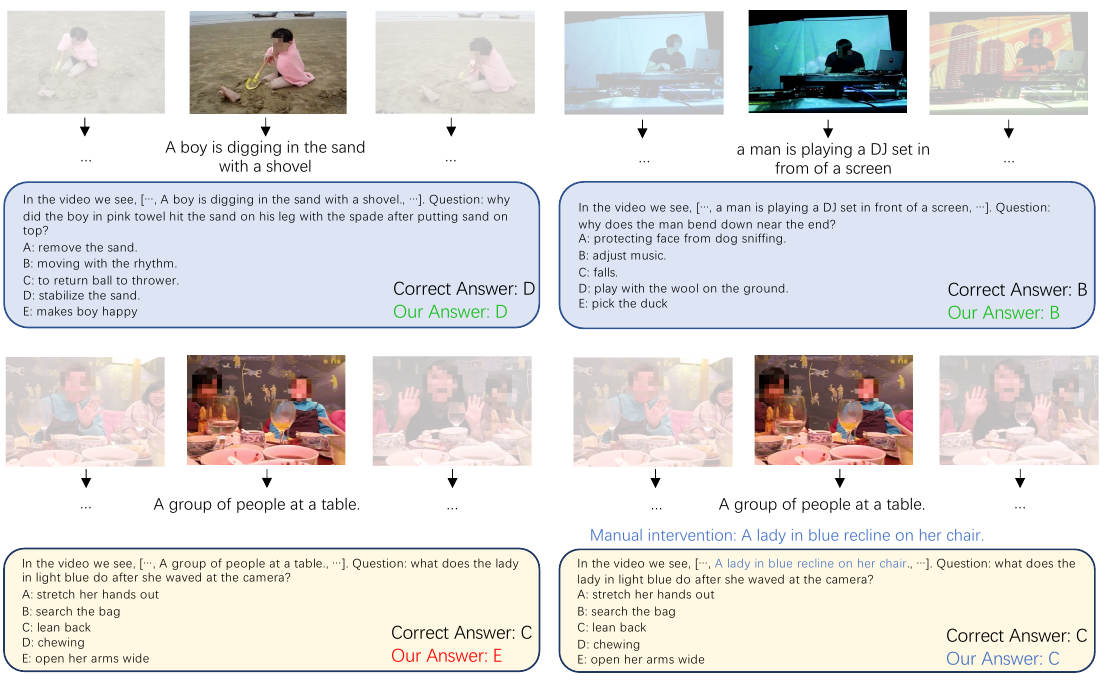}
    \vspace{-0.2in}
    \caption{More examples of Vamos video question answering and manual intervention.}
    \label{fig:intent_vis_2}
\vspace{-0.2in}
\end{figure*}

\begin{figure*}[h]
    \centering
    \includegraphics[width=\textwidth]{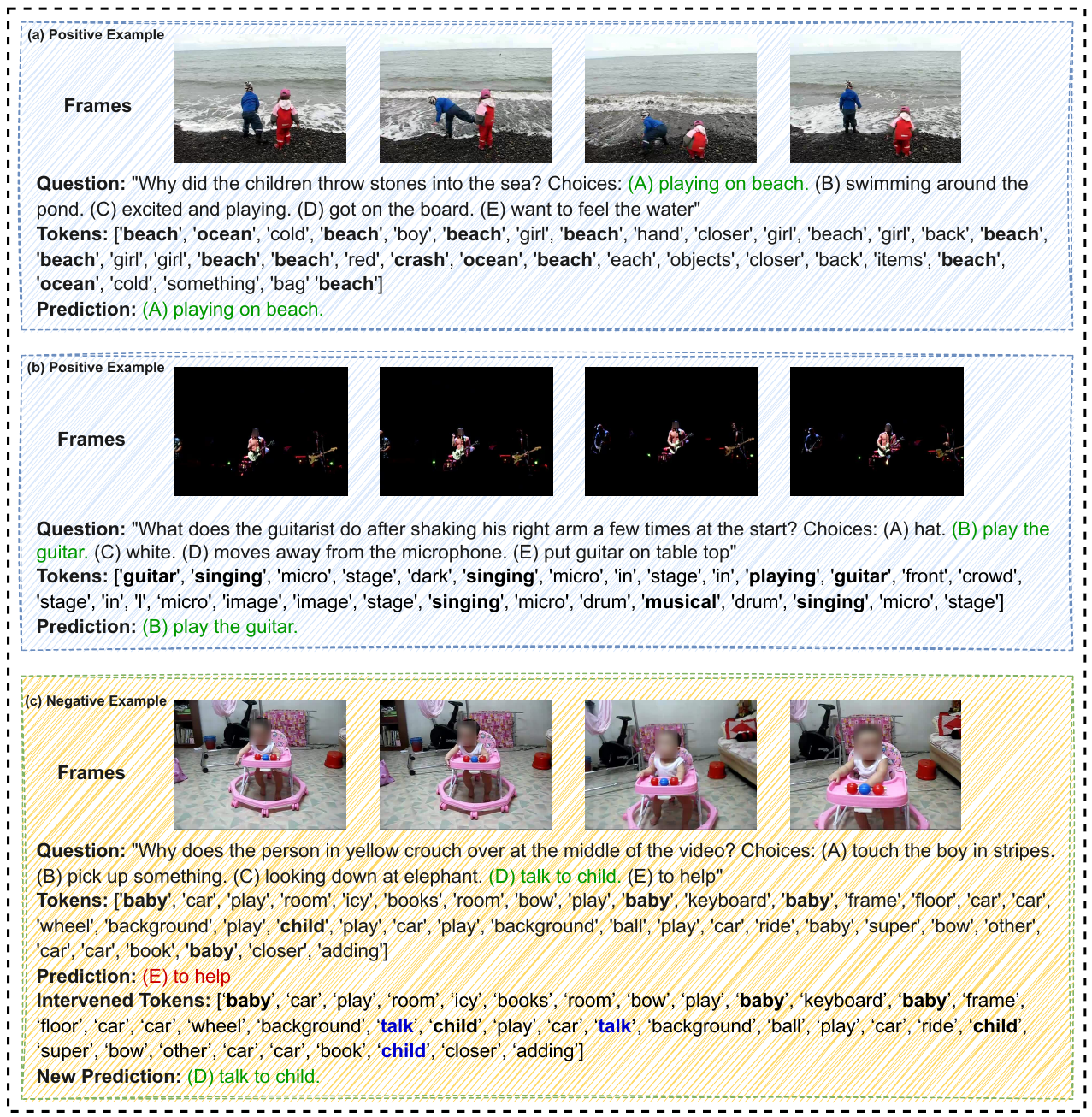}
    \vspace{-0.2in}
    \caption{Additional illustration of predictions with TBM and manual intervention. The token bottleneck selects tokens highly related to the question and intervention corrects wrong predictions without training.}
    \label{fig:apt_vis_2}
\vspace{-0.2in}
\end{figure*}

\begin{figure*}[h]
    \centering
    \includegraphics[width=\textwidth]{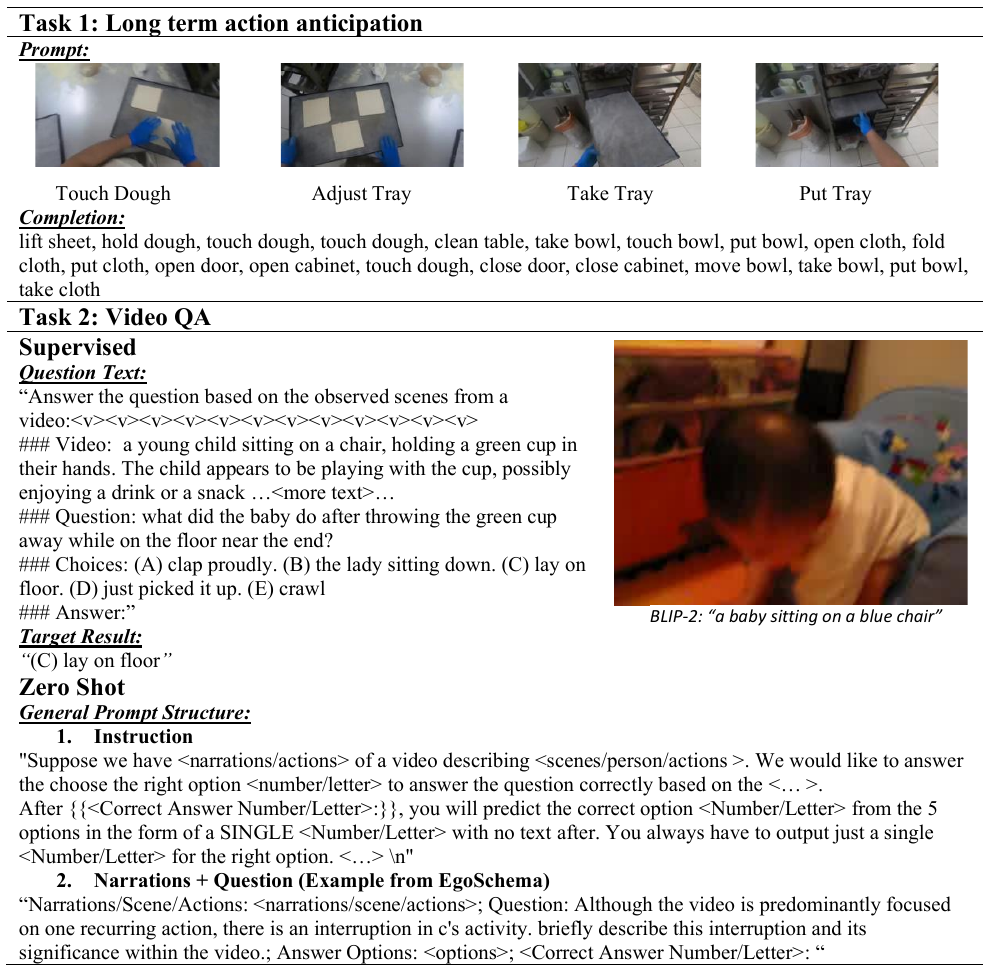}
    \vspace{-0.2in}
    \caption{Input of Vamos for long-term action anticipation and video question answering. \texttt{<v>} in prompts denotes vision tokens.}
    \label{fig:taskvisshort}
\end{figure*}

\begin{figure*}[h]
    \centering
    \includegraphics[width=\textwidth]{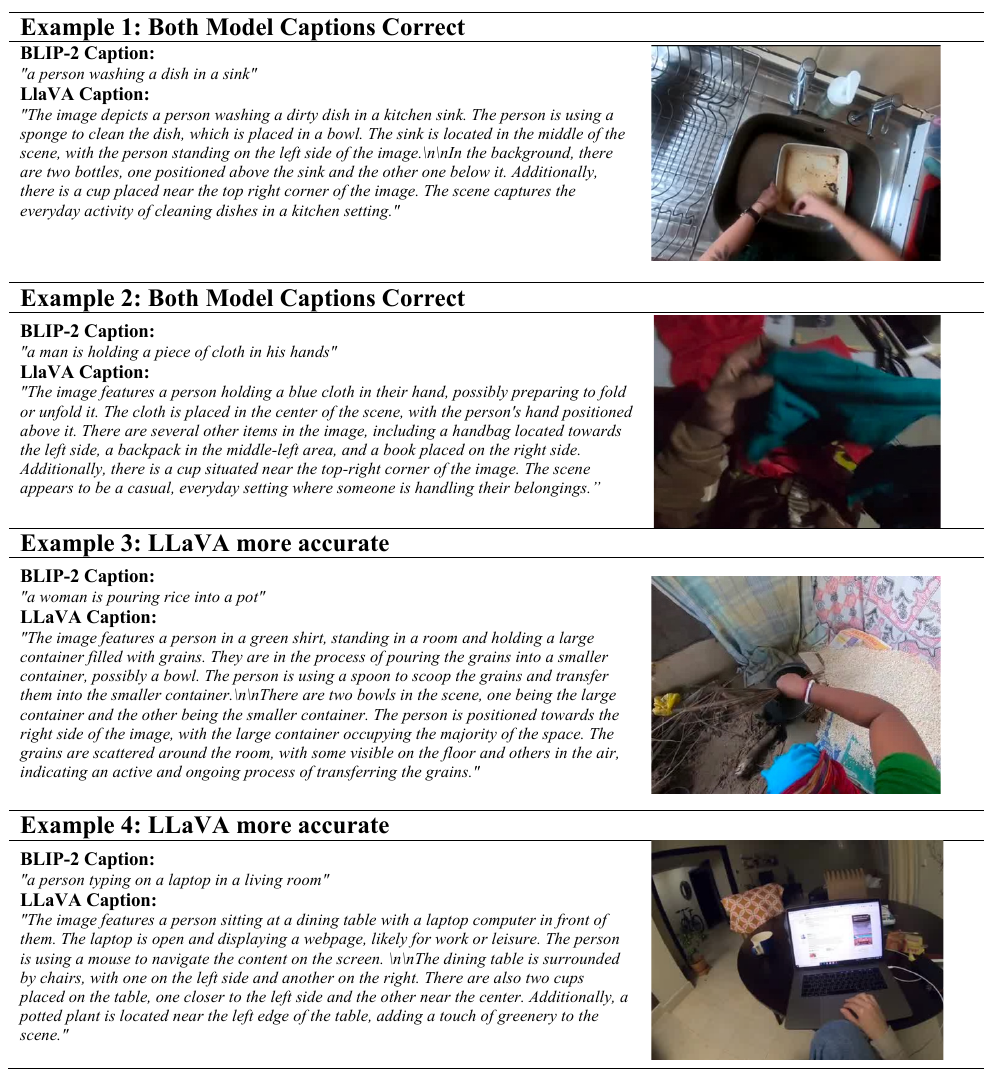}
    \vspace{-0.2in}
    \caption{Example of video captions generated by BLIP-2 and LLaVA on the NeXT-QA dataset, respectively.}
    \label{fig:cap_vis}
\end{figure*}

In Figure~\ref{fig:intent_vis_2} and Figure~\ref{fig:apt_vis_2}, we provide additional prediction examples from Vamos and Vamos with the token bottleneck model respectively.

\end{document}